\pdfoutput=1

\documentclass[11pt]{article}
\usepackage{acl}

\usepackage{times}
\usepackage{latexsym}
\expandafter\def\expandafter\UrlBreaks\expandafter{\UrlBreaks
  \do\a\do\b\do\c\do\d\do\e\do\f\do\g\do\h\do\i\do\j%
  \do\k\do\l\do\m\do\n\do\o\do\p\do\q\do\r\do\s\do\t%
  \do\u\do\v\do\w\do\x\do\y\do\z\do\A\do\B\do\C\do\D%
  \do\E\do\F\do\G\do\H\do\I\do\J\do\K\do\L\do\M\do\N%
  \do\O\do\P\do\Q\do\R\do\S\do\T\do\U\do\V\do\W\do\X%
  \do\Y\do\Z}
\usepackage[T1]{fontenc}
\usepackage{amsmath}
\usepackage{cleveref}
\usepackage{booktabs} 
\usepackage{colortbl} 
\usepackage{graphicx}
\usepackage{authblk}
\usepackage{listings}
\usepackage{algpseudocode}
\usepackage[utf8]{inputenc}

\usepackage{microtype}
\usepackage{graphicx}
\usepackage{algorithm}
\usepackage{algpseudocode}
\usepackage{inconsolata}
\usepackage[toc,page,header]{appendix}
\usepackage{minitoc}

\usepackage{enumitem}

\usepackage{graphicx}

\usepackage{soul}
\definecolor{darkgreen}{HTML}{38761d}
\definecolor{darkyellow}{HTML}{ffa500}
\definecolor{LimeGreen}{RGB}{179,253,148}
\definecolor{DarkLimeGreen}{RGB}{124,176,102}
\definecolor{DarkPink}{RGB}{255,135,135}
\definecolor{OliveGreen}{RGB}{0,0.6,0}
\definecolor{hupo}{RGB}{202,105,36}
\definecolor{blue}{RGB}{0,191,243}

\DeclareRobustCommand{\hlpink}[1]{{\sethlcolor{pink}\hl{#1}}}
\DeclareRobustCommand{\hlgreen}[1]{{\sethlcolor{LimeGreen}\hl{#1}}}
\DeclareRobustCommand{\hlyellow}[1]{{\sethlcolor{yellow}\hl{#1}}}

\DeclareRobustCommand{\hlhupo}[1]{{\sethlcolor{hupo}\hl{#1}}}
\DeclareRobustCommand{\hlblue}[1]{{\sethlcolor{blue}\hl{#1}}}

\newcommand{\stitle}[1]{\vspace{1ex} \noindent{\bf #1.}}

\crefformat{section}{\S#2#1#3}
\crefformat{subsection}{\S#2#1#3}
\crefformat{subsubsection}{\S#2#1#3}
\crefrangeformat{section}{\S\S#3#1#4 to~#5#2#6}
\crefmultiformat{section}{\S\S#2#1#3}{ and~#2#1#3}{, #2#1#3}{ and~#2#1#3}

\Crefformat{figure}{#2Fig.~#1#3}
\Crefmultiformat{figure}{Figs.~#2#1#3}{ and~#2#1#3}{, #2#1#3}{ and~#2#1#3}
\Crefformat{table}{#2Tab.~#1#3}
\Crefmultiformat{table}{Tabs.~#2#1#3}{ and~#2#1#3}{, #2#1#3}{ and~#2#1#3}
\Crefformat{appendix}{#2Appx.~\S#1#3}
\crefformat{algorithm}{Alg.~#2#1#3}

\title{DiscoSum: Discourse-aware News Summarization}

\author[1]{\bf Alexander Spangher\thanks{Equal contribution.}}
\author[1]{\bf Tenghao Huang$^{*}$}
\author[2]{\bf Jialiang Gu$^{*}$}
\author[3]{\bf Jiatong Shi}
\author[4]{\bf Muhao Chen}
\affil[1]{University of Southern California Information Sciences Institute}
\affil[2]{School of Computer Science, Wuhan University}
\affil[3]{Independent Contributor}
\affil[4]{University of California, Davis}
\affil[$$]{\texttt{spangher@usc.edu}}

\begin{document}
\maketitle
\doparttoc 
\setcounter{parttocdepth}{2} 
\faketableofcontents 

\begin{abstract}
Recent advances in text summarization have predominantly leveraged large language models to generate concise summaries. However, language models often do not maintain long-term discourse structure, especially in news articles, where organizational flow significantly influences reader engagement. We introduce a novel approach to integrating discourse structure into summarization processes, focusing specifically on news articles across various media. We present a novel summarization dataset where news articles are summarized multiple times in different ways across different social media platforms (e.g. LinkedIn, Facebook, etc.). We develop a novel \textit{news discourse schema} to describe summarization structures and a novel algorithm, \textbf{DiscoSum}, which employs beam search technique for structure-aware summarization, enabling the transformation of news stories to meet different stylistic and structural demands. Both human and automatic evaluation results demonstrate the efficacy of our approach in maintaining narrative fidelity and meeting structural requirements.
\end{abstract}

\section{Introduction}

In recent years, text summarization has seen remarkable advances, fueled by foundational Large Language Models that produce concise, context-rich overviews of lengthy documents {\cite{li-chaturvedi-2024-rationale, peper2024pelms,zhang-etal-2024-benchmarking}. Yet despite these gains, current summarization approaches rarely account for a fundamental aspect of textual organization: discourse structure \cite{cohan-etal-2018-discourse}. 


\begin{figure}[t]
    \centering
    \small
    \includegraphics[width=.9\linewidth]{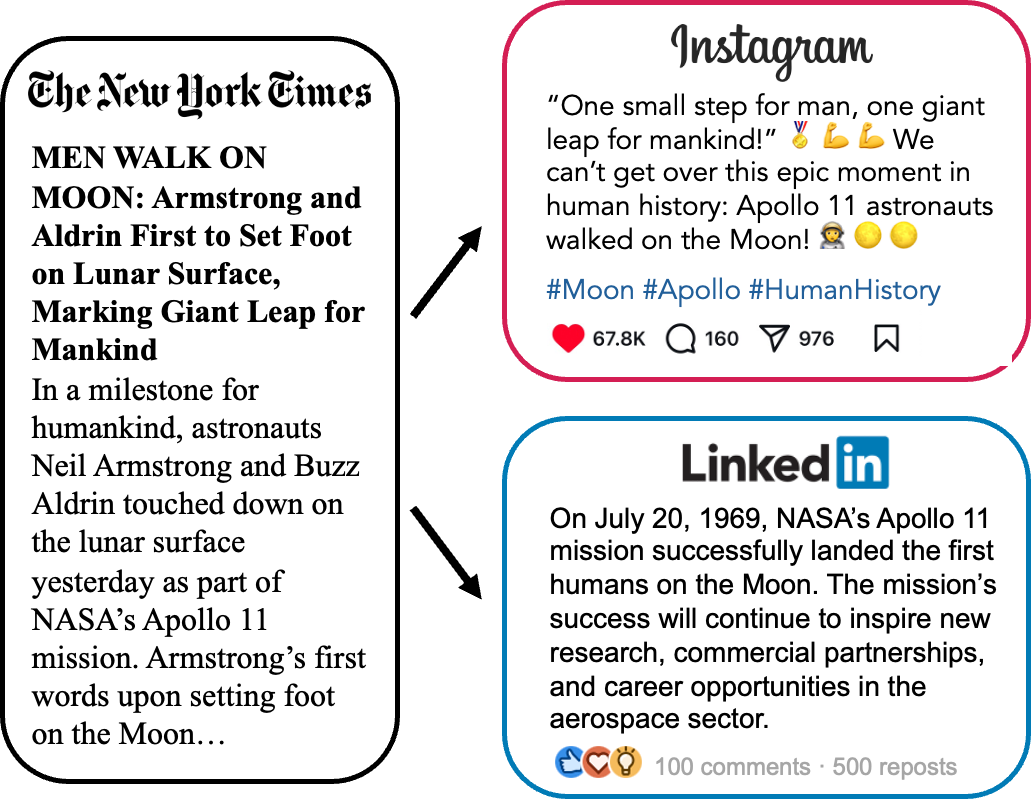}
    \caption{Comparative presentation of the Apollo 11 moon landing news across multiple platforms by The New York Times. This example showcases the diversity in content formatting and language adaptation for different audiences: a detailed traditional print article, a concise and visually-driven Instagram post, and a professionally oriented LinkedIn summary. Each platform reflects specific editorial strategies to engage its unique audience effectively, underlining the importance of discourse-aware news summarization.}
\label{fig: teaser_figure}
\vspace{-3mm}
\end{figure}

Modern news organizations like \textit{the New York Times} increasingly publish news summaries in a variety of media (e.g. print newspapers, mobile apps, podcasts, and social media) each with distinct audience expectations and content formats \cite{kalsnes2018understanding,ngoc2022journalism}. For instance, an outlet like The New York Times may produce a child-friendly podcast edition that uses simplified language and gentler framing, a condensed Instagram version with concise, visually engaging snippets, and a longer, more detailed write-up on LinkedIn or the newspaper's own website to cater to professional or academic readers. Transforming a single piece of news into multiple styles and lengths, while preserving its core narrative and emphasis, demands \textbf{nuanced control over discourse structure} \cite{shen2017styletransfernonparalleltext, hu2017toward}.

Despite the growing interest in automated news summarization \cite{see2017pointsummarizationpointergeneratornetworks, zhang2020pegasus, beltagy2020longformer, he2020ctrlsum, zhao-etal-2022-read}, existing dataset approaches have overlooked this need\footnote{See Appendix \ref{app:newsroom_comparison} for a deeper comparison to \citet{grusky2020newsroomdataset13million}.}. To bridge these gaps, we propose a novel discourse-structure-aware summarization task that emphasizes the modeling of structural discourse beyond surface-level summarization coherence or factual correctness. 

First, we introduce \textbf{DiscoSum}: a \textbf{Disco}urse-aware News \textbf{Sum}marization dataset. DiscoSum represents the largest and most diverse collection of professionally-written cross-platform news summaries, comprising 20k news articles from 23 different news outlets across 10 countries, multiply paired with over 100k human-written summaries from 4 distinct platforms: Facebook, Instagram, Twitter and newsletters. Next, we develop a novel discourse schema to describe structural components of news summaries, consisting of five sentence-level discourse labels.
%
%
Finally, we also propose a novel discourse-driven decoding method that employs a beam search technique to evaluate and select the optimal subsequent sentences for inclusion in summaries. We evaluate our method by developing both surface-level and structural metrics to assess the effectiveness of models in producing structure-aware summaries. Our human and automated evaluations confirm that our approach effectively maintains narrative fidelity and adheres to structural demands. In summary, we make the following contributions:
\begin{enumerate}
    \item \textbf{New Task:} We introduce structure-aware summarization into the news domain.
    \item \textbf{New Dataset:} We introduce a large-scale corpus of 20k news articles paired one-to-many with >100k different human-written summaries on Facebook, Twitter, Instagram and newsletters. We introduce a novel discourse schema for structural summarization.
    \item \textbf{Benchmark Results:} We present baseline models and evaluations demonstrating the feasibility and potential of NLP systems for improving structure-aware news summarization.
\end{enumerate}

\section{Related Works}
\stitle{News Summarization} News summarization has been a key focus of natural language processing research \cite{barzilay2005sentence, hong2014repository, paulus2017deepreinforcedmodelabstractive,goyal2022news}. Traditional methods often rely on extractive techniques, such as selecting "lead" sentences that approximate the news "lede" \cite{fabbri-etal-2019-multi, wang-etal-2020-heterogeneous}, but recent advancements in neural abstractive models have enabled more coherent and contextually rich summaries \cite{li-chaturvedi-2024-rationale, peper2024pelms,zhang-etal-2024-benchmarking}. Large-scale datasets, including those specifically curated for news articles, have further propelled model performance by providing diverse and representative training samples \cite{grusky2020newsroomdataset13million, chen2016thorough}. However, many of these approaches do not explicitly model the news article's inherent structure leading to summaries that, while fluent, may omit crucial structural components \cite{grenander-etal-2019-countering, zhao-etal-2022-read}.

\stitle{Controllable Generation and Test-Time Alignment} Controllable generation has emerged as a promising way to ensure outputs satisfy certain style, tone, or length requirements \cite{yang2019controllable, yang-klein-2021-fudge, Zhao2022RevisitingGC}. 
One notable area of research within controllable generation is \textit{test-time alignment}, where models incorporate constraints or preferences at inference time to better conform to user or task-specific guidelines. Techniques such as prompt engineering or decoding-time gating have shown promise in guiding model outputs toward desired attributes \cite{ Meng2022ControllableTG, Huang2023AffectiveAD, Liu2024MonotonicPI}. However, these methods often focus on surface-level constraints—like word count or style—and may not account for the deeper discourse structures characteristic of news articles.

\stitle{Discourse-Aware Language Modeling} A growing body of work highlights the significance of discourse structures—such as identifying a document's truning points, sources, and concluding remarks—in improving text generation tasks \cite{Zhai2003BeyondIR, tian-etal-2024-large-language,  spangher-etal-2024-explaining, spangher-etal-2025-creative}. Discourse-aware methods leverage theories like discourse elements \cite{spangher-etal-2022-sequentially} or journalistic guidelines \cite{spangher-etal-2022-newsedits} to parse and utilize the structural components of text during generation. While some efforts incorporate rhetorical roles or discourse parsing in domain-specific tasks \cite{wang-cardie-2013-domain, wang-ling-2016-neural}, their application to news articles is still nascent. By aligning with the natural organization of journalistic text, these methods show promise for generating summaries that both inform and engage, bridging the gap between factual coherence and audience-oriented design.

Our work differs from several prior attempts at structured summarization in other domains. The STRONG framework \cite{zhong2023strong} uses structure to parse legal documents to determine which elements to include in summaries, rather than controlling the generation structure itself. Similarly, work on dialogue summarization \cite{chen2023controllable} employs structural controls such as entity tuples and dialogue act distributions but focuses on local coherence rather than global structure. Most similar to our approach is research on meta-review summarization \cite{shen2022mred}, though it relies on hand-crafted and manually labeled articles with one-to-one article-to-summary mappings. Our work introduces methods for structured summarization without hand-labeling and allows for one-to-many mappings.

\section{Task and Dataset}
In this section, we describe the task formulation and evaluation metrics of structural summarization (\cref{ssec:task_formulation}). We introduce our proposed dataset including its composition and annotation process (\cref{sec:dataset}). 

\subsection{Task Formulation}
\label{ssec:task_formulation}
Let \(D\) denote the original news document, which can consist of multiple paragraphs or sentences. We define a desired sequence of discourse labels as \(\mathbf{T} = (t_{1}, t_{2}, \dots, t_{n})\), where each \(t_{i}\) represents a discourse label (for instance, ``contextual details,'' or ``introductory elements,'' etc.) that the \(i\)-th sentence of the summary should fulfill. The objective is to generate a summary \(\mathbf{S} = (s_{1}, s_{2}, \dots, s_{m})\), where each \(s_{i}\) is a sentence relevant to \(D\) and coherent. 
\textit{\ul{Note:}} In the \emph{structured summarization} task we assume that the user supplies the target label sequence \(\mathbf{T}\) \emph{a~priori}\footnote{This mirrors real newsroom workflows where social--media editors routinely apply pre--defined templates for different platforms.  For example, commercial content–automation systems such as \textit{Automated Insights} populate fixed headline and body layouts, and studies in discourse analysis show that canonical forms recur across news \cite{van1988news,dai2018finegrained} and even classical essay writing \cite{montaigne1580essays}.}. \emph{Predicting} an optimal structure for new input is left for future work. 

We employ a classification function \(C(\cdot)\) that, given a sentence, predicts its discourse label. Let \(\mathbf{L} = (l_{1}, l_{2}, \dots, l_{m})\) be the sequence of labels predicted by \(C(s_{i})\) for every sentence \(s_{i}\) in \(\mathbf{S}\). We require \(\mathbf{L}\) to align with \(\mathbf{T}\) in order, so that \(l_{i} = t_{i}\) for each position \(i\). Although the most straightforward scenario sets \(m = n\), such that the summary contains exactly \(n\) sentences, more flexible variants may allow for slight deviations while still ensuring that core positions match the targeted labels.


\subsection{Dataset}
\label{sec:dataset}\label{ssec: dataset}
We seek to construct a large, diverse dataset of news articles matched with multiple different summaries of each article, written by journalists, across different social media platforms and newsletters. We collect a list of 23 different major national and international news outlets\footnote{The New York Times, 
The Wall Street Journal, 
Washington Post, 
AP News, 
BBC, 
Reuters, 
The Guardian, 
Bloomberg, 
Times of India, 
Le Monde, 
The New Zurich Times, 
El País, 
China Daily, 
Los Angeles Times, 
Chicago Tribune, 
The Boston Globe, 
USA Today, 
The Sydney Morning Herald, 
The Japan News, 
De Zeit} from 10 different countries (U.S., China, India, U.K., Germany, etc.), in order to capture a range of different discourse styles across different writing styles.

\begin{table}[t]
  \centering
  \small
  \renewcommand{\arraystretch}{1.0} 
  \setlength{\tabcolsep}{18pt} 
  \begin{tabular}{l r} 
    \hline
    \textbf{Category} & \textbf{Count} \\ 
    \hline
    \# of Outlets & 23 \\
    \# of News Articles & 20,811 \\
    \# of Facebook Posts & 18,275 \\
    \# of Instagram Posts & 66,030 \\
    \# of Twitter Posts & 8,977 \\
    \# of Newsletters & 10,506 \\
    \hline
  \end{tabular}
  \caption{Overall counts of different categories.}
  \label{tab: overall_counts_categories}
\end{table}

\begin{table}
\centering
\small
\begin{tabular}{lr}
\hline
\textbf{Types} & \textbf{Counts} \\ 
\hline
\textbf{Overall} & 45,195 \\
News Article → Tweet & 12,516 \\
News Article → Facebook Post & 15,645 \\
News Article → Instagram Post & 7,738 \\
News Article → Newsletter Post & 9,296 \\
\hline
\end{tabular}
\caption{Statistics on the news article to summary graph, showing the number of edges between post types.}
\label{tab: dataset_statistics}
\end{table}

\smallskip\noindent\textbf{Social Media Collection}
We collect two years of social media posts on Twitter, Facebook and Instagram from each of the 23 news outlets. To do so, we build semi-automated scrolling agents that scroll down the feed of each news outlet's media page. We collect the full HTML of each post, including the text of each post as well as any linked urls. In total, we collect 8,977 Twitter posts, 18,275 Facebook posts, and 66,030 Instagram posts (see \Cref{tab: dataset_statistics} for more details). In order to identify structural summaries, we further filter these posts down to posts that contain 50 or more characters. This eliminates around 30\% of our data.

\smallskip\noindent\textbf{Newsletter Collection}
We select 7 newsletter brands published by news outlets,\footnote{\textit{Axios} ``The Finish Line'';
\textit{the New York Times}, ``The Morning'', 
\textit{the LA Times}, 	``California Today'';
\textit{The Skimm}, ``The Daily Skimm'';
\textit{The Daily Beast}, ``Cheat Sheet'';
\textit{Semafor}, ``Newsletters'';
\textit{CNN},	``Reliable Sources''} specifically searching for those that make all past newsletters within each brand available online in archives. We build scrapers to collect full HTML of each newsletter and collect 2 years worth of data, or over 20,000 newsletters (see Table \Cref{tab: overall_counts_categories} for details).

A newsletter often summarizes many news articles at the same time, yet our task is a single-document summarization task. Hence, we need to parse the text of each newsletter so that blocks of newsletter text correspond to single news article. This is \textit{text segmentation with overlapping segments}, since links in newsletters might require larger text segments. To accomplish this, we prompted LLMs\footnote{Prompts shown in Appendix \ref{app:newsletter_chunking}.}, building off prior work demonstrating LLM effectiveness for text segmentation tasks \cite{nayakdoes,zhao2024meta,fan2024uncovering,jiang2023advancing}. We selected a prompt configuration that instructs an LLM to (1) identify all news content links, (2) extract the surrounding text context for each link, (3) exclude boilerplate content, and (4) maintain the exact original text. To mitigate potential biases or hallucinations, we implemented a verification procedure where the largest extracted blocks are cross-checked against the LLM's own outputs in multiple iterations, with any inconsistencies flagged for manual review. Manual inspection confirmed the LLM's capability in this task, with segmentation quality exceeding 95\% accuracy in our audits across a randomly sampled set of 100 newsletters. In total, we generate 10,506 summaries from the newsletters we collect.

\smallskip\noindent\textbf{News Article Collection}
We collect a superset of news article URLs from all the social media posts and newsletters described above. Following \citet{spangher2024llms}, we scrape Wayback Machine for the HTML of each news article. We use an LLM (GPT-4) to clean the HTML to extract a full, complete news article (we find that existing libraries\footnote{https://newspaper4k.readthedocs.io/en/latest/} are insufficient). Our prompting strategy instructs the model to filter out non-news segments (e.g., login prompts, advertisements, and extraneous content), while retaining only article content. 


\smallskip\noindent\textbf{News Article and Summary Matching}
For many social media posts, we have a URL in the post that gives us an explicit match; however, for others we do not (e.g. Instagram does not allow URLs in posts). To discover as many edges as possible, we decide to match \textit{any} news article from \textit{any} outlet with \textit{any} social media post or newsletter summary.
To do so, we employ a two-step rank-and-check method. Specifically, we first use SBERT \cite{reimers2019sentence} to embed news articles and summaries; for each news article, we found the 10 closest summaries as candidates. Then, we use GPT-4 to perform a strict pairwise comparison for each candidate, returning only binary "yes" or "no" judgments on whether they describe the same news story, following the methodology validated in \citet{spangher-etal-2024-tracking}\footnote{Authors found that LLMs could be used to verify cross-document event coreference with high performance.}. In manual audits, this matching step exceeds 95\% accuracy, demonstrating the robustness of our multi-step procedure. Not only does this approach help us recover all summaries produced by a single news outlet for each article they publish, but we can see how \textit{other} news outlets cover the same news event.

\smallskip\noindent\textbf{Dataset Splits}
For all experiments, we use a 70\%/20\%/10\% train/validation/test (14k/4k/2k article-summary pairs) split of the DiscoSum dataset. This split is made at the article level to prevent leakage, so all summaries of the same article are kept within the same split. 


\section{Method}
\label{sec:method}
In this section, we outline our methods for generating structure-aware summaries. First we describe two necessary components: (1) the discourse schema we use to drive structural summarization, and (2) a sentence-level labeler, that predicts discourse labels, which we use to guide generations (\cref{ssec:discourse_schema}-\cref{ssec: discriminator}). Then, we propose two algorithms to generate summaries conforming to a target discourse sequence \(\mathbf{T}\): (1) an edit-based approach (\cref{ssec:edit_based}) and (2) a beam search method (\cref{ssec:sentence_level_beam_search}).

\subsection{Discourse Schema Generation}\label{ssec:discourse_schema}

To formalize a notion of ``structured'' summaries, we seek to construct a low-dimensional, novel discourse schema to describe social media and newsletter summaries.
First, we use an automated process to generate a schema, in contrast to prior work using manual analysis to develop schemas, typically based on O(10) examples\footnote{For example, \citet{van1988news} builds their schema based on an analysis of 12 news articles.}. Inspired by \citet{pham2024topicgpt}, we first ask an LLM to generate descriptive labels for the discourse role of each sentence in all of our summaries (O(100k) sentences). Then, we embed these labels using an SBERT embedding model \cite{reimers2019sentence}, and cluster these embeddings using k-means. 

From this embedding process, we identify five distinct clusters that represent different narrative roles: \hlpink{Introductory Elements}, \hlhupo{Contextual Details}, \hlgreen{Event Narration}, \hlblue{Source Attribution} and \hlyellow{Engagement Directive}). See \Cref{tab:label_definitions} for  definitions of each discourse role. We confirm the validity of this schema by asking two professional journalists to assess the quality and ideate for missing role labels. The choice of specifically five discourse labels was informed by extensive experimentation. While alternative parameter choices (e.g., k=7, 13, or 23) were feasible in our clustering approach, we selected a 5-dimensional schema based on human evaluation trials that showed high inter-annotator agreement ($\kappa = 0.615$) for assessing the validity of these labels. Though a 5-dimensional schema may appear limited for capturing the full complexity of news discourse structures—particularly across cross-cultural or niche news scenarios—it provides a strong foundation for this pilot study in discourse-aware summarization. 

\subsection{Discourse Labeler}
\label{ssec: discriminator}

Next, in order to guide our structure-aware generation (Section \ref{ssec:sentence_level_beam_search}), we construct a sentence-level classifier that assigns discourse labels to sentences, following \citet{spangher2021multitask, spangher-etal-2022-sequentially}. The classifier was trained on the train split of DiscoSum. To verify the quality of the validation set, we had two expert annotators independently label a subset of 500 sentences. The trained labeler achieved a high accuracy rate of over 90\% on the validation set, as shown in \Cref{fig: confusion_matrix}. This high level of accuracy is crucial for its role in the summarization process, where it is later used as a reward guidance mechanism to ensure that generated summaries adhere to the required discourse structure. The full confusion matrix in the appendix illustrates the labeler's strong performance across all five discourse categories, with the lowest per-category F1 score still exceeding 0.85.

\subsection{Generation Methods}
\subsubsection{Iterative Editing}
\label{ssec:edit_based}

Our first strategy approaches summary generation as an iterative refinement process. We begin by prompting the LLM to produce a complete initial summary, then repeatedly ``edit'' any sentences that do not fulfill their intended discourse labels. After the initial summary is generated, we use our labeler \(C(\cdot)\) to identify which sentences carry the wrong labels. We then remove these ``mismatched'' sentences and generate new candidate sentences. Over several iterations, the summary gradually ``evolves'' to match the sequence \(\mathbf{T}\). 

By focusing only on individual problematic sentences, this approach preserves what is already correct in the summary. It can also adapt to complex label sequences without having to restart the entire generation each time a mismatch is found.

\begin{algorithm}[t]
\caption{Sentence-Level Discourse-driven Beam Search with Beam Size \( k \)}
\label{alg:beamsearch}
\begin{algorithmic}[1]
\Require Source text \( X \), target label sequence \( t_1,\dots,t_N \), beam width \( k \)
\Ensure Best summary \( S = \langle s_1, s_2, \dots, s_N \rangle \)
\State Initialize beam: \( \mathcal{B} \gets \{ [] \} \) \Comment{Start with an empty sequence}
\For{\( i \gets 1 \) to \( N \)}
    \State \( \mathcal{B}' \gets \emptyset \)
    \For{\( s \in \mathcal{B} \)}
        \State \( \mathcal{C} \gets \text{LLM}(s, X, k) \)
        \Comment{\raggedright Generate \( k \) candidate }
        \For{\( c \in \mathcal{C} \)}
            \State \( s' \gets \text{append}(s, c) \)
            \State \( \text{score} \gets C(s', t_i) \)
            \State \( \mathcal{B}' \gets \mathcal{B}' \cup \left\{ (s',\, \text{score}) \right\} \)
        \EndFor
    \EndFor
    \State \( \mathcal{B} \gets \text{selectTopK}(\mathcal{B}', k) \)
\EndFor
\State \( S \gets \operatorname{argmax}_{(s,\text{score}) \in \mathcal{B}}\, \text{score} \)

\Return \( S \)
\end{algorithmic}
\end{algorithm}

\subsubsection{Sentence-Level Beam Search}
\label{ssec:sentence_level_beam_search}

In contrast to iteratively fixing errors, our second strategy constructs a label-compliant summary sentence by sentence from scratch in a beam search style \cite{Lowerre1976TheHS}. 

We begin with an empty summary and consider one position at a time (e.g., first the sentence that should have the ``introductory elements'' label, then the sentence that should have the ``contextual details'' label, and so on). At each step \(i\), the LLM generates several candidate sentences (forming a sentence-level ``beam''), which are then evaluated by \(C(\cdot)\). We choose the candidate that best matches the target label \(t_i\). This sentence is appended to the current partial summary. By evaluating multiple options at each step and selecting the best match for the desired label, this approach ensures each summary sentence follows the intended label sequence. 
The detailed procedure is described as \cref{alg:beamsearch}.

\section{Experiments}

In this section, we present our experimental setup (\Cref{sec:implementation_details}) and evaluation framework for structured summarization with target discourse labels (\Cref{sec:eval_protocol}). We introduce baseline models and methods being benchmarked (\Cref{sec:baseline_models}). Next, we present empirical results (\Cref{sec:main_results}), human preference evaluation (\Cref{sec:human_pref_eval}) and the analysis on the impact of different beam sizes (\Cref{sec: beam_size_analysis}).

\subsection{Implementation Details}
\label{sec:implementation_details}

For vanilla generation, we sample the best output among 16 trials based on automated discourse labeler. In the Sentence-Level Beam Search, we employ \(\text{BeamSize} = 16\). We fine-tuned the LLaMa-3-8B model using the PEFT method on the train split of \textbf{DiscoSum}. This fine-tuning approach reduced the validation loss significantly over 20 epochs. Key hyperparameters included a learning rate of 5e-05 and a multi-GPU distributed training setup across eight Nvidia 4090. For each generation in our experiments, we randomly generate a list of structural tags, to simulate the widest possible set of user inputs. This also prevented us from overfitting on commonly observed discourse structures.

\subsection{Evaluation Protocols}
\label{sec:eval_protocol}

\stitle{Content Accuracy Evaluation} To quantify how the content accuracy of generated news summaries, we employ several metrics:
\begin{itemize}[leftmargin=1em,itemsep=0em]

    \item \textbf{ROUGE-L.} \cite{lin-2004-rouge}  
    ROUGE-L, originally designed for summarization, measures the longest common subsequence of tokens between the generated summary and a reference summary. 

    \item \textbf{FactCC.} \cite{kryscinski-etal-2020-evaluating}  
    FactCC is a model-based metric that classifies whether each generated sentence is factually consistent with the source document.
    
    \item \textbf{AlignScore.}
    AlignScore is a consistency metric that directly measures the factual correspondence between source and summary.

\end{itemize}

\stitle{Structural Evaluation}
To assess the alignment between the generated summary \(\mathbf{S}\) and the expected discourse structure $\mathbf{T}$, we derive a predicted label sequence \(\mathbf{L}\) from $\mathbf{S}$, formally \[
\mathbf{L} = \text{Labeler}(s_1, s_2, \ldots, s_n), s_i \in S
\]
where \(\text{Labeler}\) represents either the human annotator or the automated model designed to identify discourse structures.

We employ three metrics to quantify the closeness of \(\mathbf{L}\) to the target label sequence \(\mathbf{T}\), which represents the ideal structural roles of sentences in the summary:
\begin{itemize}[leftmargin=1em,itemsep=0em]
    \item \textbf{Longest Common Subsequence (LCS).}
    LCS measures the length of the longest subsequence common to \(\mathbf{L}\) and \(\mathbf{T}\). A higher LCS value indicates that the predicted labels closely preserve the intended label order.

    \item \textbf{Match Score.}
    The Match Score assesses the number of exact position-wise matches between \(\mathbf{L}\) and \(\mathbf{T}\). This metric reflects the precision in predicting each label at its correct position in the sequence.

    \item \textbf{Levenshtein Distance.} \cite{Levenshtein1965BinaryCC}
    This metric calculates the minimum number of single-element edits (insertions, deletions, or substitutions) required to transform \(\mathbf{L}\) into \(\mathbf{T}\). A lower Levenshtein Distance indicates a higher degree of sequence similarity.
\end{itemize}

\noindent
Given the potential high cost of human evaluation, we provide protocols for both automated and human assessments:


\stitle{Human Evaluation}
We worked with two human annotators to manually assess the discourse structure of each sentence in the generated summaries. In this study, we ask annotators to evaluate 100 summaries for each model.

\subsection{Baselines}
\label{sec:baseline_models}
To evaluate the effectiveness of our proposed approach, we benchmark it against a range of baseline models that vary in architecture, training paradigms, and optimization goals. These models include both proprietary systems and open-source alternatives, providing a comprehensive overview of current state-of-the-art capabilities in text summarization and related tasks.

\stitle{Close-source LLMs} These models, such as DeepSeek-V3 \footnote{https://api-docs.deepseek.com/news/news1226}, Claude-3-5-sonnet \footnote{https://www.anthropic.com/claude/sonnet}, and GPT-4o \footnote{https://openai.com/index/hello-gpt-4o/}, are included primarily to help us gauge how well our approach performs in comparison to leading-edge technology, even if these models are not the primary focus of our evaluation.

\stitle{Open-Source LLMs} Models like Qwen-2.5 and various configurations of LLaMa-3-8B represent more accessible options that are widely used in academic research. Each variant of LLaMa-3-8B—whether it be the vanilla version, edit-based modifications, or fine-tuned iterations—serves to illustrate different potential improvements and trade-offs within the open-source framework.

\begin{table*}[ht]
\centering
\setlength{\tabcolsep}{8pt}
\resizebox{0.9\textwidth}{!}{%
\begin{tabular}{l c c c c c c c c c}
\toprule
 & \multicolumn{3}{c}{Content Accuracy} & \multicolumn{3}{c}{Auto Struct.} & \multicolumn{3}{c}{Human Struct.}\\ 
\cmidrule(lr){2-4} \cmidrule(lr){5-7} \cmidrule(lr){8-10}
\textbf{Models} & \textbf{R-L (\%)} $\uparrow$ & \textbf{FactCC} $\uparrow$ & \textbf{AlignScore} $\uparrow$ & \textbf{MS} $\uparrow$ & \textbf{Lev} $\downarrow$ & \textbf{LCS} $\uparrow$  & \textbf{MS} $\uparrow$ & \textbf{Lev} $\downarrow$ & \textbf{LCS} $\uparrow$\\ 
\midrule
\rowcolor[HTML]{EFEFEF} \multicolumn{10}{c}{\textbf{Proprietary Models}} \\ 
\midrule
DeepSeek-V3   & \underline{47.15}  & 0.47 & 0.3886  & 0.26  & 0.64  & 0.65 & 0.24	&	0.65 &	0.65\\
Claude     & 34.30  & \textbf{0.70} & 0.3882  & 0.25  & 0.68  & 0.64 & 0.20 &		0.49	& 0.75 \\
GPT-4o     & 29.51  & 0.63 & 0.3884  & 0.11  & 0.80  & 0.62 & 0.15	&	0.58 &	0.68 \\ 
O1 & 44.65 & 0.50 & - & 0.28 & 0.66 & 0.54 & - & - & -\\
\midrule
\rowcolor[HTML]{EFEFEF} \multicolumn{10}{c}{\textbf{Open-sourced Models}} \\ 
\midrule
Qwen-2.5         & 40.82  & 0.58 & 0.3888  & 0.24  & 0.66  & \underline{0.65} & 0.15	 &	0.52 &	0.64 \\
LLaMa-3-8B   & \textbf{47.18} & 0.50 & 0.3496  & 0.21  & 0.77  & 0.36 & 0.24 & \underline{0.49} & 0.65 \\
\quad -- Finetuned   & 22.01  & 0.61 & 0.3495  & 0.14  & 0.77  & 0.45 & 0.18 & 0.55 & \underline{0.72} \\
\quad -- Edit-based  & 15.28  & 0.59 & -  & \underline{0.51}  & \underline{0.48}  & 0.56 & \underline{0.24}	&	0.65	& 0.36 \\
\quad -- Beam Search  & 42.98  & \underline{0.64} & \textbf{0.3890} & \textbf{0.72}  & \textbf{0.32}  & \textbf{0.68} & \textbf{0.55} & \textbf{0.17} & \textbf{0.87} \\
\bottomrule
\end{tabular}%
}
\caption{Comparison of models on various metrics. Metrics are categorized into content accuracy and structural assessments, both automated and human-annotated. The metrics include ROUGE-L (\%), FactCC, AlignScore (for factual consistency), Match Score (MS), Levenshtein Distance (Lev), and Longest Common Subsequence (LCS). $\uparrow$ for higher is better and $\downarrow$ for lower is better. Boldfaced numbers highlight the best performance, while underscored numbers denote notable but secondary performances in each category.}
\label{tab:main_result}
\end{table*}

\subsection{Main Results}
\label{sec:main_results}

\stitle{Content Accuracy Evaluation} Table~\ref{tab:main_result} shows both surface-level and structural evaluations for a variety of models. Despite fluctuations in ROUGE-L, FactCC, and AlignScore across different systems, our approach—specifically the beam search variant of LLaMa-3-8B—maintains competitive performance in surface-level metrics. Notably, our beam search method achieves the highest AlignScore (0.3890), demonstrating superior factual consistency with source documents compared to both proprietary and other open-source models. This is particularly significant as it shows that structural improvements can be achieved without sacrificing—and in fact can enhance—factual alignment with source content. We also include the reasoning--centric model \textit{O1}, which outperforms GPT-4o on several metrics yet still lags behind our LLaMa-3-8B beam--search variant.

\stitle{Structural Evaluation} 
Significantly, our approach excels in both automatic and manual structural evaluations, where it demonstrates notable enhancements over both open-source baselines and the more sophisticated proprietary models. The beam search variant of LLaMa-3-8B consistently aligns more closely with the designated discourse label sequences, evidenced by its superior Match Score and reduced Levenshtein Distance. This enhancement in structural alignment underscores the model's ability to adhere rigorously to specified rhetorical structures without significant loss in surface-level accuracy. By achieving an effective balance between textual overlap and structural fidelity, our method significantly enhances the controllability and coherence of generated text.

\stitle{Performances of Edit-based and Finetuned Methods} The edit-based method demonstrates a promising capability in enhancing the structural alignment of generated summaries with the desired discourse labels, as evidenced by its strong performance in structural evaluations. However, this structural fidelity comes at a cost to the content accuracy and fluency, where the ROUGE-L scores considerably lower than other methods. This decline indicates that while the edit-based approach effectively molds the structure of the summaries, it may deviate significantly from the original text's semantic and syntactic properties.

The finetuned variant of the LLaMa-3-8B model, on the other hand, shows a less impressive adaptation to the task. Despite the potential for finetuning to tailor model behavior closely to specific datasets or task requirements, the observed performance metrics suggest a failure to capture the deeper, structural nuances necessary for this specific discourse-driven summarization task. The low scores imply that mere finetuning may be insufficient for tasks that require a deep understanding and transformation of text according to complex labeling schemes. This underperformance highlights the need for more advanced approaches to our task.

\subsection{Human Evaluation of Summary Quality}
\label{sec:human_pref_eval}

\begin{figure}[t]
    \centering
    \small
    \includegraphics[width=.7\linewidth]{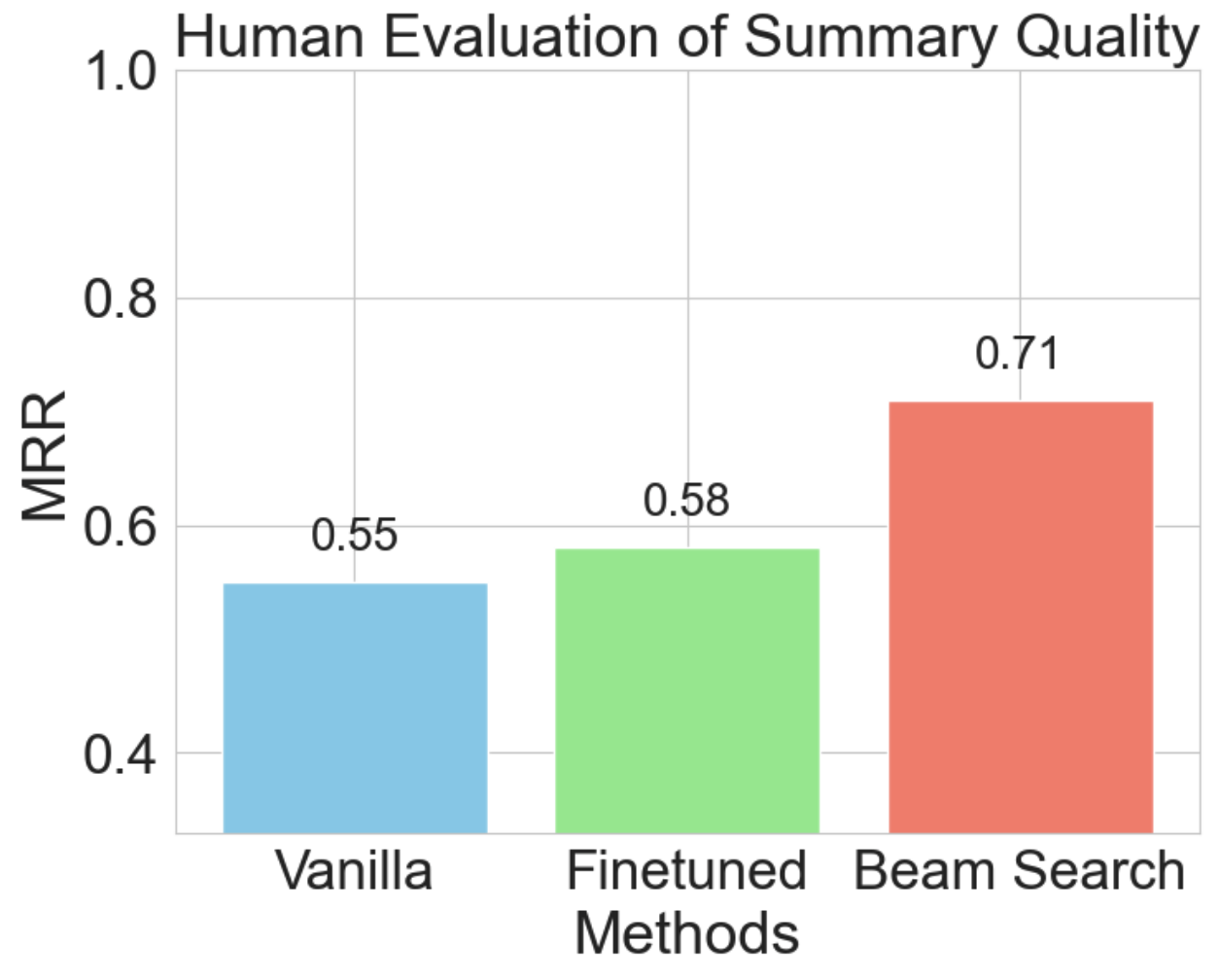}
    \caption{ Mean Reciprocal Rank (MRR) scores from human preference evaluations of summary quality across three methods: Vanilla LLaMa-3-8B, Fine-tuned LLaMa-3-8B, and Beam Search LLaMa-3-8B. }
\label{fig: human_eval_mrr}
\vspace{-3mm}
\end{figure}
We recruited two annotators to ranked the summaries based on content accuracy and structural adherence for three summary generation methods—Vanilla LLaMA-3-8B, its fine-tuned counterpart, and our beam search method. Our results, depicted in Figure \ref{fig: human_eval_mrr}, demonstrate a significant superiority of the beam search method, achieving a mean reciprocal rank (MRR) of 0.71, compared to 0.55 and 0.58 for the Vanilla and fine-tuned methods, respectively. 

\subsection{The Impact of Beam Size}
\label{sec: beam_size_analysis}
Our analysis incorporated a range of beam sizes from 2 to 16. As the beam size increases, we observe an overall improvement in the LCS scores, indicating enhanced alignment with the target discourse structure. Conversely, the Levenshtein Distance, which measures the edit distance necessary to align the predicted sequence with the target, exhibits a general decrease as the beam size increases, suggesting that larger beam sizes improve structural alignment.

The observed trends open several avenues for future research. One potential area is the exploration of adaptive beam sizes that could dynamically adjust based on the complexity of the text or the specific requirements of the discourse structure at different points in a document. Additionally, while beam search techniques enhance the quality and relevance of summaries during the inference time, integrating these high-quality summaries during training could potentially elevate the model's overall performance. Future research could look into harnessing these refined outputs to boost the training process.

\begin{figure}[t]
    \centering
    \includegraphics[width=.8\linewidth]{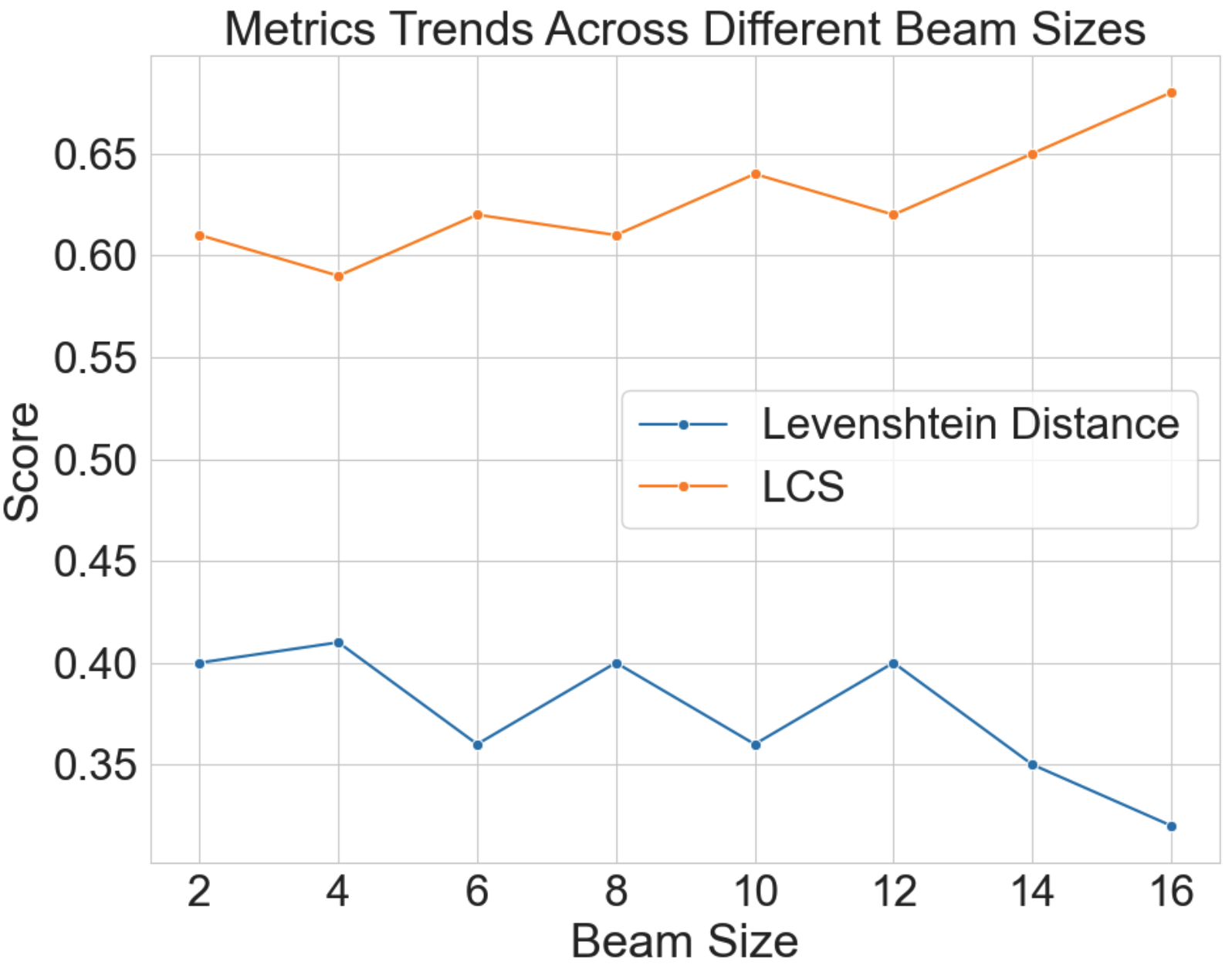}
    \caption{Levenshtein Distance and Longest Common Subsequence (LCS) scores as a function of beam size in structured summarization. The graph shows a general decrease in Levenshtein Distance and a gradual increase in LCS scores, indicating improved structural alignment with larger beam sizes.}
\label{fig:beam}
\vspace{-3mm}
\end{figure}

\section{Conclusion}
In this study, we introduced a structural summarization approach that integrates discourse organization into the summarization of news articles, emphasizing narrative fidelity and structural alignment. Our novel dataset, DiscoSum, and evaluation metrics underscore the effectiveness of our methods, particularly the beam search technique, which ensures summaries are both contextually relevant and structurally precise. The results demonstrate significant improvements over traditional methods, suggesting that our approach can enhance automated news summarization across diverse media platforms.

Our contributions highlight unexplored research problems in the field of news summarization. Our \textbf{DiscoSum} dataset and corresponding evaluation metrics set the foundation for further exploration into how discourse elements can be systematically incorporated into summarization models. This shift towards a deeper understanding of discourse structures not only challenges existing models but also opens pathways for more sophisticated approaches to news narrative reconstruction. By emphasizing structuring over surface-level coherence, we invite the research community to explore novel methodologies that could change how news content is summarized across diverse media landscapes \cite{caswell2018automated, spangher-etal-2022-newsedits, caswell2024telling, welsh2024newshomepageshomepagelayoutscapture}.

\section*{Limitations}
\stitle{Focus of the Study}
Although we measure content accuracy using standard metrics (e.g., FactCC, ROUGE-L) and acknowledge its importance, our primary goal is to ensure structural alignment with discourse labels rather than to optimize factual correctness. Consequently, improvements in factual precision or content coverage are incidental rather than intentional. Future work could investigate techniques that integrate more robust fact-checking and retrieval-augmented generation to complement structural fidelity, particularly in applications where factual accuracy is critical.

\stitle{Trade-offs in Decoding Efficiency} While our beam search method significantly improves structural adherence, it can be more computationally expensive compared to simpler generation techniques. This overhead may pose a challenge for real-time applications or large-scale deployment. Future research could explore adaptive beam strategies or hybrid methods that balance decoding speed with the need for strict discourse control.

\stitle{Potential Data Biases} Our data collection methodology involves LLMs for several critical tasks, including HTML cleaning, newsletter segmentation, and article-summary matching. While we have taken extensive steps to validate these processes, these models may introduce biases that affect the dataset's composition and the resulting schema. To mitigate this concern, we collected a diverse dataset spanning 23 major news outlets from 10 different countries across 4 different distribution methods, which helps balance potential biases across different writing styles and outlet preferences. 

Additionally, while our discourse schema is intentionally coarse-grained to enhance generalizability, we acknowledge that biases in structures can still occur. Although our primary focus was on structural rather than lexical aspects, entity or gender biases identified in prior work \cite{spangher2024llms} could potentially percolate to structural patterns. The size and diversity of our dataset help mitigate these concerns, but future work should explore the relationship between lexical biases and discourse structures, particularly for applications that require cross-cultural or domain-specific adaptations.


\bibliography{custom}

\clearpage
\onecolumn
\appendix
\addcontentsline{toc}{section}{Appendix}  
\part{Appendix}
\parttoc                                   
\twocolumn

\section{Discourse Schema Definition}

As stated previously, our discourse schema seeks to capture the structural organization of news content across different platforms and formats. The schema was developed to capture common discourse elements that serve narrative roles - from establishing context and introducing key events to providing background details and engaging readers. By modeling these discourse roles, we enable summarization systems to maintain the essential structural components that make news content coherent and engaging, rather than focusing solely on surface-level linguistic features. See Table \ref{tab:label_definitions} for a complete list of schema elements and their definitions.

The discourse labels in our schema capture high-level functional roles that sentences play within the broader narrative structure of news content. These include elements such as \hlgreen{Event Narration} (describing primary news events), \hlhupo{Contextual Details} (providing necessary background information), \hlpink{Introductory Elements} (setting up the story context), and \hlyellow{Engagement Directives} (elements designed to capture reader attention). Each label represents a distinct communicative purpose that contributes to the overall coherence and effectiveness of news storytelling, allowing our system to understand not just what information to include, but how different pieces of information function within the narrative structure.

\begin{table*}
\centering
\begin{tabular}{p{4cm}p{10cm}}
\toprule
Label & Definition \\
\midrule
\hlpink{Introductory Elements} & Sets the stage for the summary by introducing the main topic, themes, or key points that will be covered. \\
\hlhupo{Contextual Details} & Provides additional background and setting information to help understand the main events or topics being summarized. \\
\hlyellow{Engagement Directive} & Directs the reader's attention or actions through calls to action, questions, or direct addresses to engage them with the content. \\
\hlgreen{Event Narration} & Describes specific events or occurrences in a narrative form, detailing what happened in a sequential or explanatory manner. \\

\hlblue{Source Attribution} & Cites the origins of the information, giving credit to sources or clarifying the basis of the claims made in the summary. \\
\bottomrule
\end{tabular}
\caption{Discourse Label Definitions for Structured Summarization}
\label{tab:label_definitions}
\vspace{-3mm}
\end{table*}


\section{Confusion Matrix of Discourse Labeler}

In this section, we further details about the accuracy of our discourse labeler. Specifically, we present the confusion matrix of our trained discourse Labeler, compared with ground-truth annotated labels. The overall accuracy is 90.90\% and F-1 score is 0.9087. The results demonstrate the robustness our trained discourse labeler.

\begin{figure}[h]
    \centering
    \small
,l.    \includegraphics[width=\linewidth]{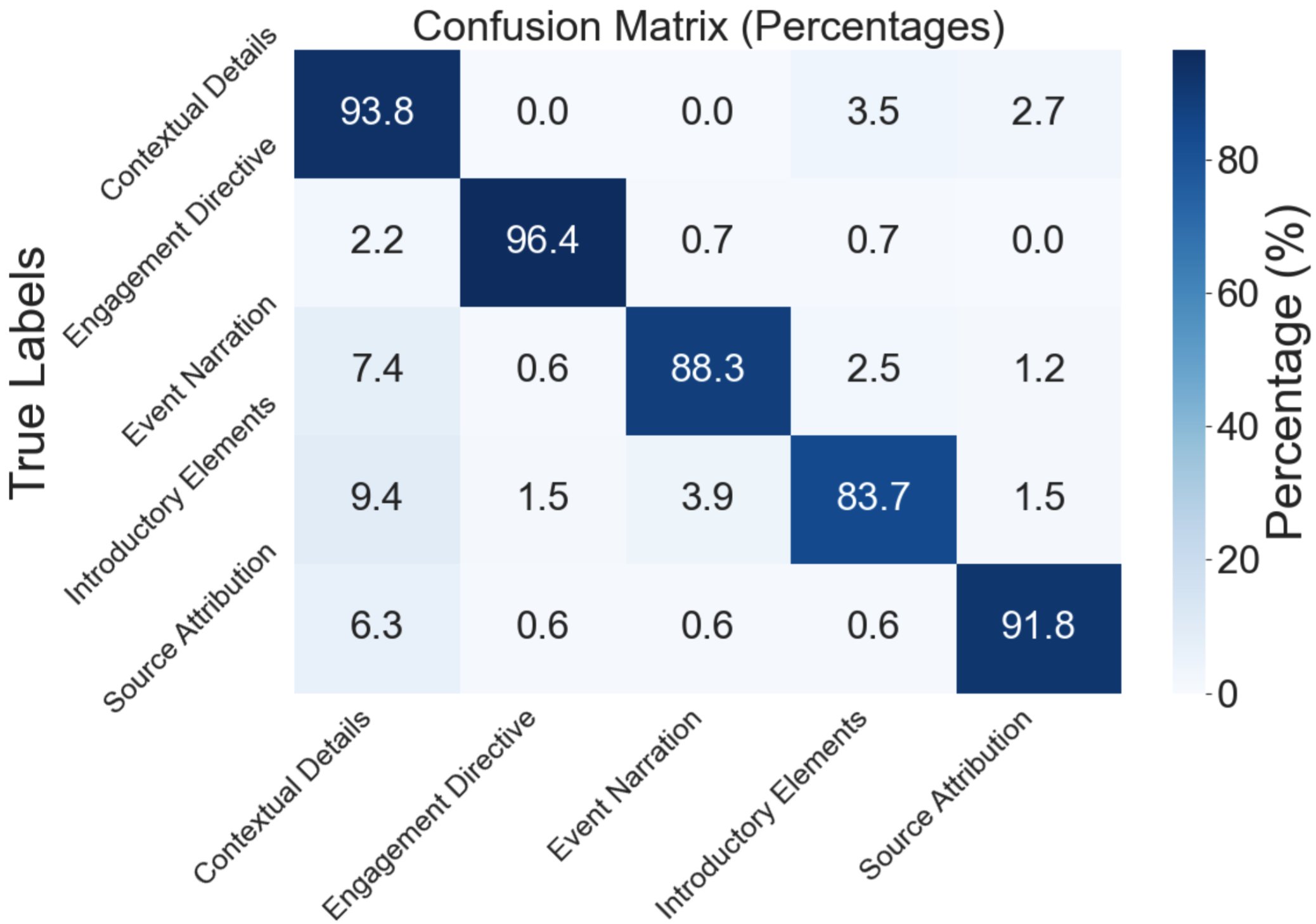}
    \caption{Confusion Matrix of Discourse Labeler.}
\label{fig: confusion_matrix}
\end{figure}

\section{Comparison to Newsroom}
\label{app:newsroom_comparison}
The NEWSROOM dataset \cite{grusky2020newsroomdataset13million} is a widely used resource for summarization research that, like our work, contains news articles and their summaries. However, our DiscoSum dataset differs fundamentally in its collection methodology, content richness, and research focus.

\subsection{Collection Methodology}
NEWSROOM's collection mechanism extracts summaries from HTML metadata, specifically the \verb|<meta property="description">...</meta>| tags embedded in article URLs. This approach efficiently collects a large volume of summaries but is limited to a single summary per article that was intended for search engine or link preview contexts.

In contrast, DiscoSum collects actual posts written by journalists for specific social media platforms and newsletters. Modern newsrooms typically employ dedicated social media teams that craft platform-specific content, resulting in multiple distinct summaries of the same article across different channels. These summaries are rarely exposed in the article's HTML metadata, as they are written directly into each platform's publishing interface.

\subsection{Content Comparison}
To illustrate this difference, we present a case study of a New York Times article about uniquely shaped Yankees baseball bats:

\begin{table*}[h]
\small
\centering
\begin{tabular}{p{2.5cm}p{9.5cm}}
\toprule
\textbf{Source} & \textbf{Content} \\
\midrule
\textbf{Article URL} & https://www.nytimes.com/athletic/6241862/2025/03/30/yankees-bats-aaron-leanhardt-marlins/ \\
\midrule
\textbf{Meta Description} \newline (NEWSROOM) & ``Aaron Leanhardt was the Yankees lead analyst in 2024 before joining the Marlins coaching staff this offseason.'' \\
\midrule
\textbf{Facebook Post} \newline (DiscoSum) & ``The New York Yankees' uniquely shaped bats have caught the attention of many and are the result of two years of research and experimentation.'' \\
\midrule
\textbf{Twitter Post} \newline (DiscoSum) & ``From @TheAthletic: The New York Yankees' uniquely shaped bats have caught the attention of many and are the result of two years of research and experimentation. Meet the former MIT physicist behind the 'torpedo' bats.'' \\
\midrule
\textbf{Instagram Post} \newline (DiscoSum) & ``The New York Yankees' uniquely shaped bat is the result of two years of research and experimentation with a former MIT physicist-turned-coach at the helm. Aaron Leanhardt, the brains behind the 'torpedo bats' making headlines, says the idea behind his innovation was simple — redistribute the weight of the bat to where it matters. The bats have been around for more than just this season. Players used them in 2024. But after last weekend's laser show in the Bronx, they have broken into the mainstream. 'Ultimately, it just takes people asking the right questions and being willing to be forward-thinking,' Leanhardt says.'' \\
\bottomrule
\end{tabular}
\caption{Comparison of content collected by NEWSROOM versus DiscoSum for the same New York Times article. Notice how different the formats are for different social media platforms.}
\label{tab:newsroom_comparison}
\end{table*}

As demonstrated in Table \ref{tab:newsroom_comparison}, the meta description (collected by NEWSROOM) is brief and focuses narrowly on the analyst's career move. In contrast, the social media posts (collected by DiscoSum) provide richer information about the story's core elements—the innovative bat design, the physics behind it, and quotes from the creator—with varying levels of detail across platforms.

\subsection{Research Value}
DiscoSum offers several advantages for summarization research:

\begin{enumerate}
    \item \textbf{Multiple reference summaries per article:} DiscoSum provides multiple professionally written summaries for each article, spanning different platforms and formats.
    
    \item \textbf{Platform-specific structural patterns:} The dataset captures how the same content is adapted for different platforms (Twitter, Facebook, Instagram, newsletters), revealing platform-specific structural patterns.
    
    \item \textbf{Real-world audience targeting:} The summaries in DiscoSum represent actual content seen by users, written by professional journalists with specific audience and platform considerations in mind.
    
    \item \textbf{Discourse structure analysis:} By annotating these varied summaries with discourse labels, DiscoSum enables research into how narrative structures adapt across platforms.
\end{enumerate}

While NEWSROOM has been invaluable for general summarization research, DiscoSum specifically enables the study of discourse-aware summarization strategies that can adapt to different platforms and structural requirements—a capability increasingly important as news consumption fragments across diverse digital channels.

\section{Prompts}

\subsection{Leaf Node Prompt}
This prompt is used to analyze and categorize discourse roles in news articles by summarizing common patterns across a set of annotated sentences. It generates concise labels that capture the essential discourse function of a group of sentences.

\begin{lstlisting}
You are a helpful assistant. I will give you a large set of notes about sentences in news articles that I wrote down.

Here are the notes:

{labels}

Please summarize them, focusing on the common discourse role each sentence plays, based on the notes. Ignore the topic.
Summarize them with a single, specific label for the entire group, being sure to concisely capture what they are about. 
Make the label 2-3 words, max. Be descriptive but not too broad. Please return just one label and one description. 
Make it in this format: ```"Label": Description```
\end{lstlisting}

\subsection{Middle Tree Prompt}
This prompt is designed for hierarchical categorization of writing elements. It helps create mid-level labels that group similar discourse roles together, focusing on common functional aspects while ignoring specific topics.

\begin{lstlisting}
You are a helpful assistant. I will give you a notes about different writing elements.

Here are the notes:

{labels}

Please summarize them with one label, focusing on the common discourse role each element plays, ignoring the topic.
Summarize them with a single, specific label for the entire group, concisely capturing what they are about. 
Make the label 2-3 words, max. Be descriptive but not too broad. Please return just one label and one description. 
Make it in this format: ```"Label": Description```
\end{lstlisting}

\subsection{Few-Shot Example Selection Prompt}
This prompt is used to identify representative examples for each discourse label. It selects diverse, high-quality examples that best illustrate a particular label, which can later be used for few-shot learning or annotation guidelines.

\begin{lstlisting}
I am trying to find good examples to use for demonstrating a label.
Here is the label: {label}. The definition for the label is: {definition}.

Here are a large set of examples I have, alone with notes for each one:
[Start Examples]
{examples}
[End Examples]

Some examples are bad. Please choose 4 examples that best represent this label. Try to pick diverse ones. 
Return the examples and the notes, and copy them fully. 
Return as a json. Be careful to format the quotes correctly.
\end{lstlisting}

\subsection{Definitions Prompt}
This prompt assigns predefined discourse role labels to sentences within social media posts. It uses contextual information from both the full post and specific annotations to match sentences with the most appropriate discourse label from a controlled vocabulary.

\begin{lstlisting}
I will give you a social media post and a single sentence from that post.
Your goal is to assign a label to that sentence with a general discourse role that best describes it's purpose in the overall script.
Each sentence also includes some notes I took about the very specific discourse role it plays, you can use them if it's helpful.

Choose from this list:
{discourse_labels}

Do NOT return any labels NOT in that list. Here are some shortened examples:
```{examples}```

Now it's your turn. Here is a social media post:
```{full_document}```

What discourse role is this sentence in it serving?
Sentence: ```{sentence}```
Notes: ```{notes}```
Answer:
\end{lstlisting}

\subsection{Newsletter Processing Prompt}
\label{app:newsletter_chunking}
This prompt extracts and organizes news content from newsletters. It identifies and separates text blocks associated with specific links, focusing on meaningful news content while filtering out boilerplate text.

\begin{lstlisting}
Look at the clean HTML of this newsletter. 

Please separate the blocks of text into news content corresponding to each individual link. 
This includes all the context surrounding the links. 
Exclude links that do not pertain to news content. 
The same text can be included in different chunks if it is relevant to a link. 
Try to include all text in at least one chunk. 
If a line doesn't end with a period, please add one. 
Do not change the text otherwise, in any way. 
Ignore text that is boilerplate and not related to news content.

Return a python dictionary mapping the link to each chunk of text. Don't return anything else. Copy the text exactly.

    ```{html}``` 
\end{lstlisting}

\section{Post Examples}
In this section, we present posts examples with discourse labels.

\begin{table*}
\centering
\begin{tabular}{p{4cm}p{10cm}}
\toprule
Label & sentence \\
\midrule
\hlpink{Introductory Elements} & Boston's streets are changing.  \\
\hlhupo{Contextual Details} & A growing number of them have bike lanes meant to protect bicyclists, slow down drivers, reduce the risk of crashes, and ultimately get more people to feel comfortable biking \\
\hlpink{Introductory Elements} & The city is aiming to expand the bike lane network so that half of residents live within a 3-minute walk of a safe and connected bike route by the end of next year. \\
\hlhupo{Contextual Details} & The theory is that if there is a safe path for biking, more people will take it, in turn reducing climate change-causing emissions, traffic deaths, and mind-numbing congestion.  \\

\hlyellow{Engagement Directive} & But challenges remain. \\
\hlpink{Introductory Elements} & Many projects face vocal opposition to ceding valuable street real estate to bikes.\\
\hlpink{Introductory Elements} & And other issues, such as the prevalence of large trucks, and lingering gaps in the bike network, make biking more dangerous than most would like. \\
\bottomrule
\end{tabular}
\caption{An example of Instagram post with sentence-level labels}
\label{tab:post_example}
\end{table*}


\begin{table*}
\centering
\begin{tabular}{p{4cm}p{10cm}}
\toprule
Label & sentence \\
\midrule
\hlpink{Introductory Elements} & Global upheaval has once again increased America's geopolitical importance.   \\
\hlgreen{Event Narration} & This years election campaign will shape the direction of U.S. policy. \\
\hlhupo{Contextual Details} & It is thus being closely watched around the world. \\
\bottomrule
\end{tabular}
\caption{An example of facebook post with sentence-level labels}
\label{tab:post_example}
\end{table*}


\begin{table*}
\centering
\begin{tabular}{p{4cm}p{10cm}}
\toprule
Label & sentence \\
\midrule
\hlpink{Introductory Elements} & Logan Edra, a 21-year-old American B-Girl, said the Olympics could provide young girls with a vision of the future.   \\
\hlblue{Source Attribution} & "Any type of representation is going to help people see what is possible." \\
\bottomrule
\end{tabular}
\caption{An example of twitter post with sentence-level labels}
\label{tab:post_example}
\end{table*}


\begin{table*}
\centering
\begin{tabular}{p{4cm}p{10cm}}
\toprule
Label & sentence \\
\midrule
\hlgreen{Event Narration} & Disney began laying off thousands of staff members, its second round of layoffs, to save \$5.5 billion in costs and cut 7,000 jobs.   \\
\hlhupo{Contextual Details} & Employees at ESPN, Disney Entertainment, Disney Parks, and Experiences and Products will also be affected.\\
\hlyellow{Engagement Directive} & A third round of layoffs is expected before summer.  \\
\hlpink{Introductory Elements} & Meanwhile, Insider's employees went on strike after about 10\% of its staff was laid off. \\
\hlhupo{Contextual Details} & Staffing cuts have also affected Buzzfeed, NPR, and other news organizations.\\
\bottomrule
\end{tabular}
\caption{An example of newslttter with sentence-level labels}
\label{tab:post_example}
\end{table*}

\begin{table*}[]
    \centering
    \begin{tabular}{p{4cm}p{10cm}}
        \toprule
         \multicolumn{2}{l}{Original News Article} \\
         \midrule
         \multicolumn{2}{p{14cm}}{There's overwhelming support for an age limit on the president and Congress, but it won't happen anytime soon. Polling by Insider and Morning Consult indicates that three in four Americans favor an age limit for members of Congress. More than four in 10 viewed the ages of political leaders as a "major" problem. Some lawmakers are concerned that politics has become a gerontocracy, and the country's oldest politicians have not just the incentive to stick around but the power to remain entrenched. The seniority system, a campaign-finance system that favors incumbents, gerrymandered congressional districts, and the ability to continue to seek office indefinitely are mechanisms that members of Congress are most eager to discuss. Age limits are not on the table, and Congress' youngest and oldest lawmakers seem to agree. The country's power class is also aging across the board; the current Congress is the oldest in the nation's history. At the start of the latest Congress, in January 2021, baby boomers controlled roughly 56\% of the seats in both chambers despite representing approximately just 21\% of the US population. Some lawmakers reflect on an underlying system that prioritizes tenure and seniority over other potential measures of political acumen, generating a political leadership that skews older. Term limits offer another potential solution, though that too would require going through the arduous constitutional-amendment process."} \\ 
         \midrule
         \multicolumn{2}{p{14cm}}{Target Labels: \hlyellow{Engagement Directive}, \hlhupo{Contextual Details}, \hlpink{Introductory Elements}, \hlpink{Introductory Elements}, \hlgreen{Event Narration}} \\
         \midrule
\hlyellow{Engagement Directive} & Many are calling for a constitutional amendment to set a mandatory retirement age for lawmakers, citing the need for fresh perspectives and a more diverse range of experiences in Washington. \\ 
\hlhupo{Contextual Details} & Additionally, the proposed age limit aims to address concerns that the current age distribution in Congress may hinder its ability to effectively represent the diverse needs and values of the younger generation. \\
\hlpink{Introductory Elements} & Despite the widespread support, opponents argue that imposing an age limit would violate the constitutional principle of equal protection, as it would unfairly target a specific group based on age. \\ 
\hlpink{Introductory Elements} & Despite the ongoing debate, polling suggests that three in four Americans support an age limit for members of Congress, with over four in 10 viewing the ages of political leaders as a major problem. \\ 
\hlgreen{Event Narration} & A recent poll has found that three in four Americans favor an age limit for members of Congress, with over four in 10 viewing the ages of political leaders as a "major" problem. \\ 
\bottomrule
    \end{tabular}
    \caption{Example generation \#1}
    \label{tab:my_label}
\end{table*}

\begin{table*}[]
    \centering
    \begin{tabular}{p{4cm}p{10cm}}
        \toprule
         \multicolumn{2}{l}{Original News Article} \\
         \midrule
         \multicolumn{2}{p{14cm}}{
         Are robot waiters the future? Some restaurants think so. MADISON HEIGHTS, Mich. (AP) \u2014 You may have already seen them in restaurants: waist-high machines that can greet guests, lead them to their tables, deliver food and drinks and ferry dirty dishes to the kitchen. Some have cat-like faces and even purr when you scratch their heads. But are robot waiters the future? It\u2019s a question the restaurant industry is increasingly trying to answer. Many think robot waiters are the solution to the industry\u2019s labor shortages. Sales of them have been growing rapidly in recent years, with tens of thousands now gliding through dining rooms worldwide. Dennis Reynolds, dean of the Hilton College of Global Hospitality Leadership at the University of Houston, says, There's no doubt in my mind that this is where the world is going. The school's restaurant began using a robot in December, and Reynolds says it has eased the workload for human staff and made service more efficient. However, others say robot waiters aren't much more than a gimmick that have a long way to go before they can replace humans. They can't take orders, and many restaurants have steps, outdoor patios and other physical challenges they can't adapt to. Redwood City, California-based Bear Robotics introduced its Servi robot in 2021 and expects to have 10,000 deployed by the end of this year in 44 U.S. states and overseas. Shenzen, China-based Pudu Robotics, which was founded in 2016, has deployed more than 56,000 robots worldwide. Li Zhai, owner of Noodle Topia in Madison Heights, Michigan, says the robots give human servers more time to mingle with customers, which increases tips. And customers often post videos of the robots on social media that entice others to visit. But not all chains have had success with robots. Chili's introduced a robot server named Rita in 2020 and expanded the test to 61 U.S. restaurants before abruptly halting it last August. The chain found that Rita moved too slowly and got in the way of human servers. Eventually, experts expect that a certain percentage of restaurants maybe 30\% will continue to have human servers and be considered more luxurious, while the rest will lean more heavily on robots in the kitchen and in dining rooms.} \\ 
         \midrule
         \multicolumn{2}{p{14cm}}{Target Labels: 
         \hlgreen{Event Narration}, \hlhupo{Contextual Details}, \hlgreen{Event Narration}, \hlyellow{Engagement Directive}, \hlgreen{Event Narration}}
         \\
         \midrule
\hlgreen{Event Narration} & Robot waiters are here to stay, and their presence is redefining the future of the restaurant industry.\\
\hlhupo{Contextual Details} & The innovative technology, first introduced at a Michigan restaurant, has been gaining popularity worldwide, as they bring efficiency, convenience, and a touch of novelty to the dining experience. \\ 
\hlgreen{Event Narration} & Robot waiters are increasingly being adopted by restaurants worldwide, offering benefits such as reduced labor costs, improved accuracy, and enhanced customer engagement. \\
\hlyellow{Engagement Directive} & As the technology continues to evolve, we can expect to see robot waiters becoming more sophisticated, with advanced navigation systems and the ability to learn and adapt to individual customers' preferences. \\ 
\hlgreen{Event Narration} & With the potential to transform the dining experience, robot waiters are being hailed as a game-changer in the industry, enabling restaurants to adapt to changing consumer preferences and improve overall efficiency, as well as providing a unique and entertaining experience for customers. \\ 
\bottomrule
    \end{tabular}
    \caption{Example generation \#2}
    \label{tab:my_label}
\end{table*}

\begin{table*}[]
    \centering
    \begin{tabular}{p{4cm}p{10cm}}
        \toprule
         \multicolumn{2}{l}{Original News Article} \\
         \midrule
         \multicolumn{2}{p{14cm}}{
\"Don't fool yourself\": Billions more needed to protect tropical forests, warns new report. At least \$130bn a year is needed to protect the most at-risk areas of tropical forest by the end of the decade, alongside reductions in beef and dairy consumption and government bans on deforestation, a thinktank has warned. Currently, finance to protect forests averages between \$2bn and \$3bn a year. The report estimates that eliminating the economic incentive to destroy forests for cattle ranching, agriculture and other uses would cost at least \$130bn a year. The money could come from carbon markets, wealthy governments and philanthropists, but there must also be urgent actions such as a ban on clearing forests, developing businesses that rely on standing forests and reducing demand for commodities linked to deforestation, such as palm oil, soya, beef and cocoa. Lord Turner, a former head of the CBI and ex-chair of the UK government's Committee on Climate Change, warned that governments should not delude themselves about the scale of the challenge, and that robust quantification of what you spend and what you get is much more difficult than anywhere else. Land use change is the second largest source of human greenhouse gas emissions, with deforestation accounting for about 15\% of the total. It is also a major driver of biodiversity loss and ecosystems degradation, and has continued at a relentless pace despite scientific warnings that ecosystems such as the Amazon could soon collapse.
         } \\ 
         \midrule
         \multicolumn{2}{p{14cm}}{Target Labels: 
         \hlgreen{Event Narration}, \hlgreen{Event Narration}, \hlgreen{Event Narration}, \hlhupo{Contextual Details}, \hlgreen{Event Narration}
         }
         \\
         \midrule
\hlgreen{Event Narration} & The new report emphasizes that protecting tropical forests is crucial for mitigating climate change, preserving biodiversity, and supporting local communities, but it requires a significant increase in funding to achieve these goals.  \\ 
\hlgreen{Event Narration} & A significant increase in funding of at least \$130 billion annually by the end of the decade is necessary, alongside reductions in beef and dairy consumption and government bans on deforestation, to effectively protect the most at-risk areas of tropical forest. \\ 
\hlgreen{Event Narration} & This funding increase is essential to protect the most at-risk areas of tropical forest, which are currently under threat due to economic incentives driving deforestation.  \\ 
\hlhupo{Contextual Details} & Currently, finance to protect forests averages between \$2bn and \$3bn a year, which is a tiny fraction of the estimated \$130bn needed annually. \\ 
\hlgreen{Event Narration} & Accordingly, the report warns that current funding levels are insufficient, and drastic measures are needed to address the issue. \\ 
\bottomrule
    \end{tabular}
    \caption{Example generation \#3}
    \label{tab:my_label}
\end{table*}

\begin{table}[]
    \centering
    \begin{tabular}{p{7cm}p{7cm}}
    \toprule
    Sentence & Superset Discourse Label \\
    \midrule
    Tina McMahon-Foley is now celebrating her 30th year as a naturalist for Cape Ann Whale Watch (@capeannwhalewatch).',	
    & "Main Event": This sentence directly describes a primary event, noting a milestone in Tina McMahon-Foley\'s career as a naturalist, which is the focal point of the document.', \\ 
    But the story of how she found her way to Cape Ann begins in her former home in Albany, when she worked as a young science teacher in the early 1990s.', 
    & "Previous Event": This sentence describes a specific event that occurred before Tina McMahon-Foley became a naturalist for Cape Ann Whale Watch, providing background information on how she found her way to Cape Ann.', \\ 
    She was watching television one night when she flipped to the Discovery Channel.', 
    & 	"Previous Event": This sentence describes a specific event that occurred before Tina McMahon-Foley became a naturalist for Cape Ann Whale Watch, providing context and background information on how she discovered her interest in whales.', \\ 
    "What came next -- a whale documentary, a drive to Massachusetts, a scientist's admiration of her gumption -- stick with her today.",	& "Consequence": This sentence describes a series of events that directly succeeded a previous event (watching the Discovery Channel) and had a lasting impact on the subject\'s life, shaping her current situation as a naturalist.' \\ 
    On a recent trip, she appeared just as excited to see a whale as she was the first day that scientist, Roger Payne, sent her to sea.', & "Anecdotal Event": This sentence describes a specific, personal experience of Tina McMahon-Foley that illustrates her enduring passion for whale watching, adding an emotional and relatable aspect to her story.', \\ 
    As a calf breached near the ship, she spoke into the mic to those on board: "Have you caught your breath yet?' & "Anecdotal Event": This sentence describes a specific, personal moment in Tina McMahon-Foley\'s experience as a naturalist, which is used to illustrate her enthusiasm and passion for her work, rather than to advance the main narrative of her 30-year career.', \\ 
    \bottomrule
    \end{tabular}
    \caption{An example of the superset of discourse labels that was used to make our discourse schema.}
    \label{tab:superset_discourse_examples}
\end{table}
    
\end{document}